\documentclass[sn-mathphys,Numbered]{sn-jnl}

\usepackage{rotate}
\usepackage{graphicx}%
\usepackage{hhline, multirow}%
\usepackage{amsmath,amssymb,amsfonts}%
\usepackage{mathtools}
\usepackage{amsthm}%
\usepackage[title]{appendix}%
\usepackage{xcolor}%
\usepackage{breqn}
\usepackage{manyfoot}%
\usepackage{algorithm}%
\usepackage{algpseudocode}%
\usepackage{makecell}

\DeclareMathOperator*{\argmax}{argmax}
\DeclareMathOperator*{\argmin}{argmin}
\newcommand{\etal}{\textit{et al}. }

\begin{document}

\title[Article Title]{Target Aware Network Architecture Search and Compression for Efficient Knowledge Transfer}

\author*[1]{\fnm{S.H. Shabbeer} \sur{Basha}}\email{shabbeerb@rvu.edu.in}

\author[2]{\fnm{Debapriya} \sur{Tula}}\email{debapriyatula@google.com}

\author[3]{\fnm{Sravan Kumar} \sur{Vinakota}}\email{skv3@njit.edu}

\author[4]{\fnm{Shiv Ram} \sur{Dubey}}\email{srdubey@iiita.ac.in}


\affil*[1]{\orgname{RV University}, \orgaddress{\city{Bengaluru}, \state{Karnataka}, \country{India}}}

\affil[2]{\orgname{Google Research}, \orgaddress{\city{Bangalore}, \state{Karnataka}, \country{India}}}

\affil[3]{\orgname{New Jersey Institute of Technology}, \orgaddress{\city{New Jersey}, \country{USA}}}

\affil[4]{\orgdiv{Computer Vision and Biometrics Lab}, \orgname{Indian Institute of Information Technology Allahabad}, \orgaddress{\city{Prayagraj},  \state{Uttar Pradesh}, \country{India}}}


\abstract{Transfer Learning enables Convolutional Neural Networks (CNN) to acquire knowledge from a source domain and transfer it to a target domain, where collecting large-scale annotated examples is time-consuming and expensive. Conventionally, while transferring the knowledge learned from one task to another task, the deeper layers of a pre-trained CNN are finetuned over the target dataset. However, these layers are originally designed for the source task which may be over-parameterized for the target task. Thus, finetuning these layers over the target dataset may affect the generalization ability of the CNN due to high network complexity. To tackle this problem, we propose a two-stage framework called TASCNet which enables efficient knowledge transfer. In the first stage, the configuration of the deeper layers is learned automatically and finetuned over the target dataset. Later, in the second stage, the redundant filters are pruned from the fine-tuned CNN to decrease the network's complexity for the target task while preserving the performance. This two-stage mechanism finds a compact version of the pre-trained CNN with optimal structure (number of filters in a convolutional layer, number of neurons in a dense layer, and so on) from the hypothesis space. The efficacy of the proposed method is evaluated using VGG-16, ResNet-50, and DenseNet-121 on CalTech-101, CalTech-256, and Stanford Dogs datasets. Similar to computer vision tasks, we have also conducted experiments on Movie Review Sentiment Analysis task. The proposed TASCNet reduces the computational complexity of pre-trained CNNs over the target task by reducing both trainable parameters and FLOPs which enables resource-efficient knowledge transfer. The source code is available at: \href{https://github.com/Debapriya-Tula/TASCNet}{https://github.com/Debapriya-Tula/TASCNet}.}




\maketitle 
\footnote{This paper is accepted for publication in Multimedia Systems Journal.}

\section{Introduction}\label{sec1}
Deep CNNs have evolved as widely accepted machine learning algorithms for solving many real-world problems due to their ease of hierarchical feature extraction \cite{lecun2015deep} from raw input images. However, hand-designing a CNN for a specific problem is a time-consuming task and also requires domain expertise \cite{szegedy2016rethinking,he2016deep}. In recent years, the problem of automatically learning the structure of CNNs, which is known as Neural Architecture Search (NAS) (a sub-domain of Automated Machine Learning) \cite{elsken2019neural} has gained the attention of researchers. However, the NAS methods available in the literature are focused on finding better performing CNNs for popular image datasets such as ImageNet \cite{deng2009imagenet}, CIFAR-10/100 \cite{krizhevsky2009learning} which have a large number of training images.


Transfer Learning enables the utilization of pre-trained CNNs for transferring knowledge learned from a source task to a target task \cite{yang2019deep}.
Typically, the pre-trained CNN layers are finetuned on the target dataset while transferring the knowledge from one domain to another. It is worthwhile to note that these pre-trained CNN layers were originally designed for the source task, which may not be appropriate for the target task. Moreover, re-training the pre-trained model over the target dataset may result in a high overfitting effect if the target dataset has fewer images.  

This paper introduces a two-stage optimization framework that enables efficient knowledge transfer from a source to a target task. Initially, the network hyperparameters involved in the deeper layers (the top layers) of a pre-trained CNN are learned w.r.t. the target task. Later, the CNN tuned in the first stage is utilized for pruning the redundant filters. To perform the first step i.e., to automatically tune pre-trained CNN layers w.r.t. the target dataset, we have used Bayesian Optimization \cite{frazier2018tutorial}. In the second stage, we prune the redundant filters that have similar weight values throughout the network training.

\begin{algorithm*}
\caption{Hyperparameter Tuning} 
\textbf{Inputs:} $\mathcal{M}$ (Pre-trained_CNN), $\mathcal{H}$ (hyperparameters search space), Train$_{Data}$, Val$_{Data}$, Epochs (number of epochs for training proxy CNNs). \\
\textbf{Output:} A pre-trained CNN having suitable structure for target dataset for better image classification.
\label{Eff_Auto_Tune_algorithm}
\begin{algorithmic}
 \Procedure{Hyperparameter Tuning}{} 
 \State Assume Gaussian Process (GP) prior on the objective $F$
 \State Evaluate the objective function $F$ at initial $k_0$ points 
 
 
 
 
 
$m = k_{0}$ 
 
\While{$i \in m+1, .., M$}   \Comment{examine the hyperparameter search space}

\State Using the available data (prior) update the posterior distribution on $F$
\State Select the next point $x_i$ from $\mathcal{H}$ at which the acquisition function value is maximum

 
\State Evaluate and observe $y_i = F(x_i)$
\EndWhile
  
\State \textbf{return} $x_i$ \Comment{pre-trained CNN with target-dependant structure} 

\EndProcedure

\end{algorithmic}
\end{algorithm*}

\section{Related Works}
\label{sec_related_works}
The ability of CNNs to extract hierarchical features enables them to transfer knowledge from one domain to another. Transfer Learning is a popularly employed deep learning practice for domains such as medicine \cite{shin2016deep,raghu2020eeg}, agriculture \cite{kamilaris2018deep}, medical \cite{ahuja2021deep}, and many more that have a limited amount of training data. Recently, an average-pooling-based classifier has been proposed to detect and recognize breast cancer images. The complexity of the base network (pre-trained CNN) is very high for a target task having a low number of training images when a pre-trained CNN is utilized for transfer learning.
\cite{molchanov2016pruning} proposed a new method for pruning network parameters in the transfer learning setting based on the Taylor expansion. The parameters that are less important for the target task are pruned to increase the generalization ability. A two-stage method is introduced in \cite{han2018new} to increase the generalization ability of a network using web data augmentation. Yosinski \etal
\cite{yosinski2014transferable} introduced a method to estimate the knowledge-transferring ability of each CNN layer.

Finding an optimal CNN architecture a.k.a Neural Architecture Search (NAS) for a given problem without human intervention has attracted researchers in recent years \cite{elsken2019neural}. Before the evolution of NAS methods, Machine Learning (ML) algorithms were benefited by hyperparameter optimization \cite{bergstra2011algorithms,snoek2012practical}. NAS methods require extensive computational resources along with search time for discovering desirable CNN architectures. To mention a few, a NAS method proposed by \cite{zoph2016neural} explored $12,800$ proxy (child) CNNs with the demand of 800 GPUs for 28 days. Similarly, NASNet \cite{zoph2018learning} made use of  $500$ GPUs for about $4$ days to train $20,000$ child CNNs which were explored during the architecture search. Recently, a multi-objective NAS \cite{jiang2020efficient} is introduced, whose objective is to minimize both classification error on validation data and the large number of parameters involved in the network.

CNN pruning is another line of research that is aimed at removing redundant filters/neurons from a neural network to increase its generalization ability. CNN pruning can be performed in many ways such as introducing sparsity into the network parameters \cite{chen2015compressing, han2015deep}, weight quantization-based pruning \cite{rastegari2016xnor, polino2018model}, and filter pruning \cite{singh2020leveraging, lin2020hrank,fan2021hfpq}. Especially, filter pruning is a generic method that enables network acceleration without the support of additional software and hardware. For example, the methods based on $\ell_1$-norm \cite{li2016pruning} and $\ell_1$-norm combined with capped $\ell_1$-norm are aimed at pruning convolutional filters.

To re-use the knowledge learned from a source task, transfer learning is a better choice for obtaining competing performance with less data \cite{weiss2016survey}. Re-training (finetuning) the last few layers of a pre-trained CNN over a target dataset yields comparable results over it. However, pre-trained or base CNN is originally proposed for the source data, which might not be relevant to the target data. A few attempts have been made in the literature to learn the deeper layers of a pre-trained CNN using the knowledge of a target dataset \cite{basha2021autofcl,basha2021autotune}.

In this direction, \cite{basha2021autofcl} showed that learning the configuration of the fully connected layers with the help of target data leads to better transfer learning ability. Later, \cite{basha2021autotune}, extended their previous work by proposing AutoTune which is aimed at learning suitable hyperparameters corresponding to the deeper layers of a pre-trained CNN. They have empirically shown that learning the target-aware deeper layers' configurations improves the transfer learning performance. However, they have also reported that the AutoFCL and AutoTune methods produce more trainable parameters and FLOPs for CNNs such as ResNet-50 and DenseNet-121.

\begin{figure*}
    \centering
    \includegraphics[width=\textwidth]{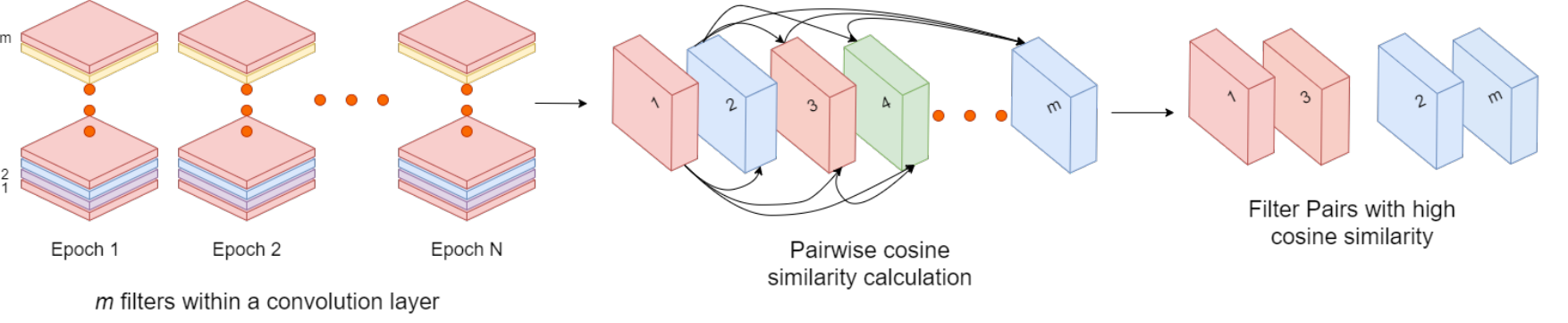}
    \caption{For each layer, cosine similarity between filters belonging to the same filter pair is computed after each training cycle. In the diagram, a cuboid, $i$ represents the collection of weights of the $i^{th}$ filter over all the training epochs.}
    \label{fig:FilterPairing}
\end{figure*}

To achieve efficient knowledge transfer from a source task to a target task, we propose a two-stage framework called TASCNet. The hyperparameters involved in the deeper pre-trained CNN layers are learned by utilizing the target dataset in the first stage. Neural network pruning not only compresses the deep neural networks but also suggests better deep networks. This motivated us to prune redundant filters throughout the CNN re-training in the next stage. The key contributions of this research are as follows,
\begin{itemize}
    \item A framework is proposed for reducing a pre-trained CNN's capacity for target task.    
    \item The hyperparameters in deeper layers of a pre-trained CNN are learned automatically.
    \item In the later stage, redundant filters are pruned to increase the model's generalizability.
    \item Our transfer learning results are compared with the state-of-the-art. 
    
\end{itemize}

\section{Proposed Method}
\label{sec:tascnet}
Tuning pre-trained CNN layers' hyperparameters automatically for better transfer learning is formulated as a Bayesian Optimization problem. Here, we consider $F$ as the objective function, and the goal of Bayesian Optimization is to find $x_*$ which is given as follows,
\begin{equation}
    x_* = \argmax_{x \in \mathcal{H}} F(x)
\end{equation}
where the input $x \in \mathbb{R}^d$ is a d-dimensional vector representing the hyperparameters and $\mathcal{H}$ is the hyperparameter search space which is shown in Table \ref{hyperparameter_space_table}. Finding the value of the objective $F$ at any point in the given hyperparameter search space is an expensive operation. This value can be found by fine-tuning the weights involved in proxy CNN layers (with the hyperparameters explored during the architecture search) on the target dataset. The CNN layers with optimal hyperparameters configuration are denoted with $x_*$. Therefore, the pre-trained CNN $\mathcal{M}$ with the target-dependent CNN structure ($x_*$) is aimed at achieving the best classification performance on the validation data. After obtaining the target-specific CNN with the optimal structure of layers, the redundant filters in the CNN are pruned to increase the model's generalization capacity. Next, we discuss how Bayesian Optimization is used to learn the target-aware CNN layers for better knowledge transfer.

\subsection{Automatically Tuning the deeper layers of a pre-trained CNN}
\label{sec:autotune}
The primary objective of this paper is to increase the network's generalization capacity over the target dataset which is achieved through two stages. In the first stage, the hyperparameters involved in the deeper layers of a pre-trained CNN are tuned automatically as depicted in Algorithm \ref{Eff_Auto_Tune_algorithm}. Later, in the second stage, the redundant filters are pruned from convolutional layers to enable efficient knowledge inference. Algorithm \ref{Eff_Auto_Tune_algorithm} considers a pre-trained CNN $M$, hyperparameter space ($\mathcal{H}$), training data (Train$_{Data}$), held-out validation data (Val$_{Data}$), and the maximum number of epochs ($Epochs$) to train each child CNN explored during the hyperparameter tuning as input for tuning the pre-trained CNN layers automatically using Bayesian Optimization \cite{frazier2018tutorial}. 

Bayesian Optimization is a popular approach adopted to tune the hyperparameters involved in deep ConvNets \cite{snoek2012practical}. Bayesian Optimization Algorithm (BOA) is used to solve problems in which each function evaluation is expensive and there is no access to the gradients. So, typically BOA is employed to solve black-box optimization problems. BOA involves two steps in modeling the objective functions. Firstly, a Bayesian statistical surrogate model is utilized which is responsible for building an approximate for the objective function $F$. In the second step, an acquisition function is involved to guide the Bayesian search by proposing the next best point. In this paper, we utilize Gaussian Processes (GP)  \cite{rasmussen2003gaussian} as the surrogate model and Expected Improvement (EI) as an acquisition function. 

\begin{figure*}[!t]
    \centering
    \includegraphics[width=\textwidth]{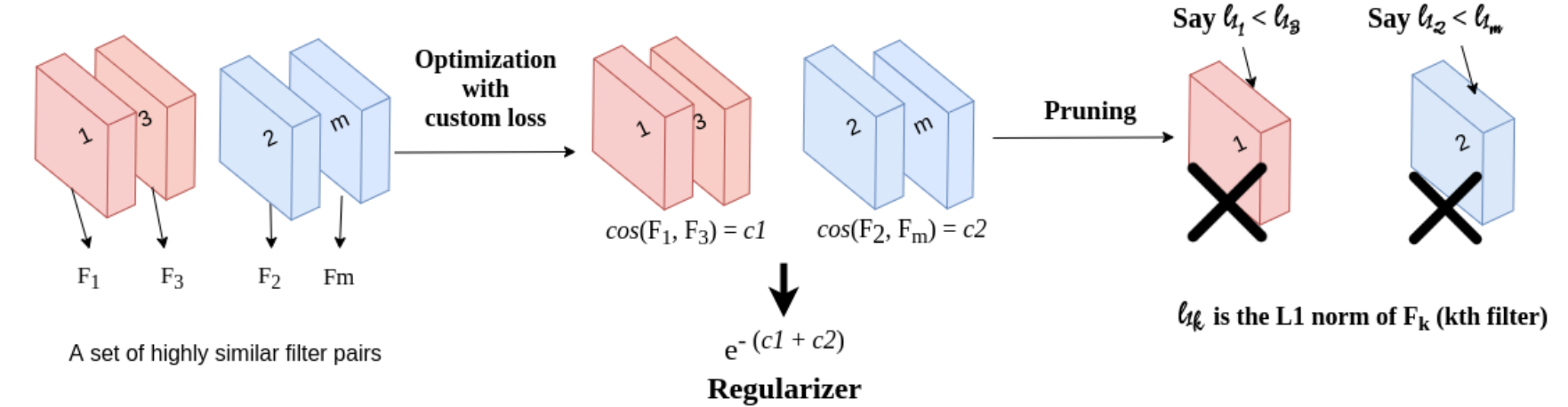}
    \caption{In each pruning iteration, the top 5\% of the filter pairs with high similarity are chosen for optimization. During optimization, we improve the similarity between the filters by re-training the model using the custom regularizer. After optimization, the filter having a lower $\ell_1$-norm compared to the other filter in the same pair will be pruned.}
    \label{fig:pruning}
\end{figure*}

Algorithm \ref{Eff_Auto_Tune_algorithm} evaluates and observes the value of the objective at initial $k_0$ points (which is set to $20$ in our experimental settings) that are chosen in a uniform manner. After observing the objective $F$ at initial $k_0$ points, an optimal CNN with relevant hyperparameter configuration $x^{+}$ is the one for which the maximum validation accuracy $F(x^{+})$ is observed while finetuning over the target dataset. Now, we have to estimate the objective $F$ at a new point $x^{1}$ i.e., $F(x^1)$. At this point, the best objective value is either the previous best i.e., $F(x^{+})$ if $F(x^+) \ge F(x^{1})$ or $F(x^{1})$ if $F(x^+) <  F(x^{1})$. Thus, the gain in the objective value is either $F(x^1) - F(x^{+})$, if $F(x^{1})$ is greater than the previous best $F(x^+)$, otherwise zero. At iteration $i=m+1$, Algorithm \ref{Eff_Auto_Tune_algorithm} chooses a new point $x^1$ at which the gain in the objective is maximum. We do not know the value of $F(x^1)$ before evaluating the objective function at $x^1$. However, computing this value for a given $x$ is expensive. Instead, we can find the value of expected improvement and select $x^1$ at which this improvement is high.

The Expected Improvement (EI) \cite{frazier2018tutorial}, popularly known acquisition function, is used to guide the Bayesian Optimization process by suggesting the next point to sample in the hyperparameter search space. The EI is represented as follows,

\begin{equation}
    EI_{k_{0}}(x) := E_{k_{0}} [[F(x^{1})-F(x^+)]^+]
\end{equation}
$E_{k_{0}}$ denotes the expectation calculated under a posterior distribution given the function $F$ evaluations at initial $k_0$ points $x_1, x_2, ... x_{k_{0}}$. The posterior probability distribution $F(x^1)$ given the previous function evaluations $F(x_{1:k_{0}})$ results in a  multi-variate normal distribution with mean $\mu_{k_{0}}(x^{1})$ and variance $\sigma^2_{k_{0}}({x^{1}})$. The same can be presented as,

\begin{equation}
    F(x^{1})|F(x_{1:k_{0}}) \sim Normal(\mu_{k_0}(x^{1}),\sigma^2_{k_{0}}(x^{1}))
    \label{conditional_bayes}
\end{equation}
here, $\mu_{k_{0}}(x^{1})$ and $\sigma^2_{k_{0}}(x^{1})$ are computed using Equations \ref{mu_formula} and \ref{var_formula}, respectively.

\begin{dmath}
\mu_{k_{0}}(x^{1})= \sum_{0}(x^{1},x_{1:k_{0}})\sum_0(x_{1:k_{0}},x_{1:k_{0}})^{-1}(F(x_{1:k_{0}})
- \mu_{0}(x_{1:k_{0}})) + \mu_0({x^{1}})
\label{mu_formula}
\end{dmath}

\begin{dmath}
\sigma^{2}_{k_{0}}(x^{1}) = \sum_0(x^{1},x^{1})-\sum_0(x^{1}, x_{1:k_{0}})\sum_0(x_{1:k_{0}},x_{1:k_{0}})^{-1}\sum_0(x_{1:k_{0}},x^{1})
\label{var_formula}
\end{dmath} 

Expected Improvement (EI) is used in line 6 of Algorithm \ref{Eff_Auto_Tune_algorithm} to choose the next point to search from the hyperparameter space that results in a maximum gain in the EI. 

\begin{equation}
    x_{i+1} = \argmax EI_{i}(x)
\end{equation}
After each function evaluation, we update the posterior distribution using Eq. \ref{conditional_bayes}.

\subsection{Pruning redundant filters}
\label{sec:pruning}
Even though the deeper layers of a pre-trained model are tuned and re-trained having access to the target dataset, the remaining pre-trained CNN layers are developed for the source task. This phenomenon limits the generalization ability of pre-trained CNN on the target task during the knowledge transfer from one task to another task due to over-parameterization. To tackle this problem, we consider the pre-trained CNN with an optimal hyperparameter configuration of CNN layers (learned in the first stage) to prune the redundant filters from the Conv layers in the next stage.

The objective of the second step is to make the CNN computationally more efficient without compromising the performance much. We can achieve this by pruning the redundant filters from the convolutional layers. The redundancy among the filters is determined using cosine similarity throughout network training. Very recently, \cite{basha2021deep} designed a filter pruning method based on the network training history. However, they have used the difference between the $\ell_1$ norms of filters as a metric to decide the redundancy between two filters. More concretely, the filter pairs having a similar $\ell_1$ norm value throughout the network training are chosen as the redundant filters. However, the filter's $\ell_1$ norm does not take the order of filter elements (parameters) into consideration \cite{basha2021deep}. So, we propose a filter pruning method that uses the Cosine similarity between two filters throughout training. The abstract view of the pruning method is shown in Fig. \ref{fig:pruning}. Our filter pruning method involves three sub-steps i) filter selection ii) optimization and iii) finetuning.
An overview of the pruning algorithm is shown in Algorithm \ref{PruningAlgo}.

\subsubsection{Filter Selection}
\label{section:FilterPairing}
In this stage, initially, we form filter pairs such that a filter in a pair has a high Cosine similarity with the other filter in the same pair. For instance, if a convolution layer has $n$ filters, we form $n \choose 2$ filter pairs. In our experiments, we consider the convolutional layers which are having more than or equal to $256$ filters ($2048$ in the case of ResNet-50) for pruning. For each convolutional layer, we calculate the similarity between the Conv filters that belong to the same filter pair. The cosine similarity between the two filters is computed as follows,

\begin{equation}
    \label{equation:CosineSimilarity}
    \cos(\pmb F_i, \pmb F_j) = \frac {\pmb F_i \cdot \pmb F_j}{||\pmb F_i|| \cdot ||\pmb F_j||}
\end{equation}

\begin{table*}[!t]
\caption{The hyperparameter search space considered for various convolutional neural network layers.}
\label{table:hyperparameter_space}
\centering
\begin{tabular}{|l|l|l|}
\hline
 \textbf{Layer type}                                                  & \textbf{Hyperparameters involved}                                                                          & \textbf{Hyperparameter values}                                                                                 \\ \hline
 Convolution                                                        & \begin{tabular}[c]{@{}l@{}}Convolution filter size\\ Stride\\ \#filters\end{tabular} & \begin{tabular}[c]{@{}l@{}}\{1, 2, 3, 5\}\\ Always 1\\ \{32, 64, 128, 256, 512\} \end{tabular}          \\ \hline
 Max-pooling                                                        & \begin{tabular}[c]{@{}l@{}}Filter size\\ Stride\end{tabular}                       & \begin{tabular}[c]{@{}l@{}}\{2, 3\}\\ Always 1\end{tabular}                                     \\ \hline
 \begin{tabular}[c]{@{}l@{}}Fully Connected or\\ Dense\end{tabular} & \begin{tabular}[c]{@{}l@{}} \#layers \\ \# neurons \\ \end{tabular}                          & \begin{tabular}[c]{@{}l@{}} \{1, 2, 3\} \\ \{64, 128, 256, 512, 1024\}\\ \end{tabular} \\ \hline
 \begin{tabular}[c]{@{}l@{}}Dropout\end{tabular} & \begin{tabular}[c]{@{}l@{}} dropout factor\end{tabular}                          & \begin{tabular}[c]{@{}l@{}} \ {[}0, 1{]} with offset 0.1\end{tabular} \\ \hline

 \begin{tabular}[c]{@{}l@{}}Activation\end{tabular} & \begin{tabular}[c]{@{}l@{}} activation-type  \end{tabular}                          & \begin{tabular}[c]{@{}l@{}} \ {Sigmoid, TanH, ReLU, ELU,  SELU}\end{tabular} \\

\hline
\end{tabular}
\label{hyperparameter_space_table}
\end{table*}

Throughout the model's training, at every epoch, the filter parameters are collected and unrolled into column vectors. These column vectors, accumulated over the training epochs are concatenated which results in a single vector $F_i$. Consider a filter of dimensions $H \times W \times C$, where $H$, $W$, and $C$ represent the height, width, and number of channels, which is trained for $N$ epochs. Then the resulting vector will be of dimension $N \times H \times W \times C$ which is further used to calculate the cosine similarity. The pictorial representation of the filter selection process is shown in Fig. \ref{fig:FilterPairing}.


If two filters have a high cosine similarity, it means that the filters represent similar information and one of them is redundant. Our method computes the cosine similarity between the two filters throughout the training which helps to identify the filters that are redundant throughout the network training. Thus, after computing the cosine similarity for all pairs of filters, we choose the top $m\%$ of filter pairs with the highest cosine similarity. In the experiments, we prune $5\%$ of filters from the convolutional layers in each pruning iteration.

\subsubsection{Optimization} 
\label{section:Optimization}

 Inspired by the two recent works \cite{singh2020leveraging,basha2021deep}, we introduce an optimization step before filter pruning to decrease the information loss that will be incurred due to filter pruning. The regularizer aims at increasing the similarity between the filters belonging to each pair, which is computed using Eq. \ref{eq:RegularizerFunction}. This also helps in increasing the generalization ability of the network. During optimization, the model is fine-tuned using the weighted cross-entropy loss function \cite{phan2020resolving} given as follows,
 
 \begin{equation}
    \label{ClassWeightedCrossEntropyLoss}
    L'(y, \hat{y}) = \sum_{k = 1}^C w_k (-y_k \log \hat{y_k})
\end{equation}
where $C$ is the number of classes, $y_k$ is ground truth and $\hat{y_k}$ is predicted label. Let $N_k$ be the number of training examples belong to the $k^{th}$ class, then the class weight $w_k$ for the $k^{th}$ class is computed as $w_k = \frac{1}{N_k}$.

The regularization term can be written as,
\begin{equation}
    \label{eq:RegularizerFunction}
    R = \exp \left(\sum_{F_i, F_j \in S}\!\!-\cos\left(F_i, F_j\right)\right)
\end{equation}
here $S$ represents the set of filter pairs having high cosine similarity (which are selected for optimization). After adding the custom regularizer (given in Eq. \ref{eq:RegularizerFunction}) to the loss function shown in Eq. \ref{ClassWeightedCrossEntropyLoss}, the new objective function is given as follows,
\begin{equation}
    \label{ModifiedLossFunction}
    W = \argmin_W \left(\sum_{k = 1}^Cw_k \left(-y_k \log \hat{y_k}\right) + R \right)
\end{equation}


\subsubsection{Pruning and Retraining:} \label{section:PruningRetraining}

The filter pairs having the high correlation are chosen in the filter selection step. After optimization, one of the filters from each pair is pruned. The filter in a pair with the least $\ell_1$-norm is added to the set of filters to be removed \cite{li2016pruning}. 
During the filter selection step, some filters may appear more than once in the selected pairs, being paired up with different filters. Thus, they may be added to the set of filters to prune more than once. As a result, in most cases, the effective number of filters considered for pruning is lesser than 5\% of the total filters in that convolutional layer.

\textbf{Exception for ResNet-50:}
ResNet-50 contains skip connections which are merged into the main branch using an \textit{Add} layer. The output feature maps of the layers used for element-wise addition should have the same dimensional feature maps. Thus, to avoid any mismatch in the depth dimension, we ensure that exactly $5\%$ of the filters are pruned from the convolutional layers (that have filters more than or equal to $2048$). 
Let $L_i^k$ be the $k^{th}$ filter in the $i^{th}$ convolutional layer and ($L_i^1$, $L_i^7$), ($L_i^1$, $L_i^4$) be two filter pairs chosen during the filter selection process. And consider $L_i^1$ has a lower $\ell_1$-norm compared to $L_i^7$ and $L_i^4$. To ensure the depth dimension compatibility, we choose $L_i^4$ from the second pair for pruning whereas $L_i^1$ is chosen from the first pair. 

\section{Search Space}
\label{sec_hyper_space}
Here, we discuss the search space used for hyperparameter tuning. The plain structured CNNs such as VGG-16 \cite{simonyan2014very} and CNNs with skip connections such as ResNet-50 \cite{he2016deep} and DenseNet-121 \cite{huang2017densely} are utilized to conduct the experiments. These models comprise various layers including convolutional layers, max-pooling layers, and dense layers. Various hyperparameters involved in these layers are tuned with the knowledge of the target dataset in the first stage of the proposed framework. It is observed from the literature that most of the CNNs \cite{lecun1998gradient,krizhevsky2012imagenet} have three fully connected layers that contain an output layer. Hence, we have chosen to search over the range \{1, 2, 3\} for the hyperparameter `number of layers'. We consider the search space as \{64, 128, 256, 512, 1024\} for the number of neurons hyperparameter. Each fully connected layer (except the output layer) is followed by a Batch Normalization layer and a Dropout \cite{srivastava2014dropout} layer. The dropout factor \cite{srivastava2014dropout} is tuned in the range \{0.1, 0.2, \ldots 0.9\}. The hyperparameter search space for other layers like convolution and max-pooling is shown in Table \ref{table:hyperparameter_space}. 



\begin{table}[!t]
    \centering
\resizebox{\textwidth}{!}{%
\begin{tabular}{|c|c|c|}
\hline
\textbf{Dataset} & \textbf{\# Images} &  \textbf{\# Classes} \\ \hline
CalTech-101 & \begin{tabular}[c]{@{}c@{}}Training: 7,316\\ Validation: 1,830\end{tabular}  & 102 \\ \hline
CalTech-256 & \begin{tabular}[c]{@{}c@{}}Training: 24,485\\ Validation: 6,122\end{tabular} &  256 \\ \hline
Stanford Dogs & \begin{tabular}[c]{@{}c@{}}Training: 16,464\\ Validation: 4,116\end{tabular} &  120 \\ \hline
The 20 News Group  & \begin{tabular}[c]{@{}c@{}}Training: 15,998\\ Validation: 3,999 \end{tabular} & 20 \\ \hline
\makecell{Movie Reviews \\ Sentiment Analysis} & \begin{tabular}[c]{@{}c@{}}Training: 28280\\ Validation: 7072\end{tabular} &  2 \\ \hline
\end{tabular}
    }
    \caption{Summary of the CalTech-101, CalTech-256, Stanford Dogs, The 20 News Group and Movie Reviews Sentiment Analysis datasets. }
\label{table:datasetsummary}
\end{table}

\begin{algorithm*}[h]
\caption{Similarity Based Filter Pruning }
\label{PruningAlgo}
\textbf{Inputs:} $\mathcal{M}$ (pre-trained CNN), Train$_{Data}$, Valid$_{Data}$, Epochs, Prune$_{percentage}$\\
\textbf{Output:} $\mathcal{M^{`}}$ (A light-weight pre-trained CNN)
\begin{algorithmic}
    \Procedure{FilterPrune}{}
    \State Load $\mathcal{M}$ (pre-trained CNN)
    \State $Curr\_Val\_Accuracy \gets \Call{Train}{\mathcal{M}}$
    \State Let $Global\_Max\_Val\_Accuracy \gets -\infty$ \Comment{Variable to keep track of the maximum validation accuracy}
    \While{$Abs(Global\_Max\_Val\_Accuracy - Curr\_Val\_Accuracy) \le Min\_Diff$}
    \If{$Global\_Max\_Val\_Accuracy \le Curr\_Val\_Accuracy$}
        \State $Global\_Max\_Val\_Accuracy \gets Curr\_Val\_Accuracy$  \Comment{Update the maximum validation accuracy}
    \EndIf
    \State $Filter\_Pairs \gets \Call{MakeFilterPairs}{\mathcal{M}}$ \Comment{Make pairs of similar filters}
    \State \Call{Optimize}{$\mathcal{M}$, $Filter\_Pairs$} \Comment{Optimize the model with custom loss}
    \State $Pruned\_Model \gets \Call{DeleteFilters}{\mathcal{M}, Filter\_Pairs}$ \Comment{Delete the redundant filters}
    \State $Curr\_Val\_Accuracy \gets \Call{Train}{Pruned\_Model}$ \Comment{Re-train the pruned model}
    \State $\mathcal{M^{`}} \gets Pruned\_Model$ \Comment{Update the model}
    \EndWhile
    \EndProcedure
\end{algorithmic}
\end{algorithm*}

\begin{table}[t]
\centering
\resizebox{\textwidth}{!}{%
\begin{tabular}{|c|c|c|c|c|c|c|c|c|c|c|}
\hline
\multirow{2}{*}{\rotatebox[origin=c]{90}{Deep CNN}} & \multirow{2}{*}{\rotatebox[origin=c]{90}{Dataset}} & \multicolumn{4}{c|}{\rotatebox[origin=c]{90}{FC Layer}} & \multicolumn{4}{c|}{\rotatebox[origin=c]{90}{Convolution Layer}} & \multirow{2}{*}{\rotatebox[origin=c]{90}{Validation Accuracy}} \\ 

                              &                                    & \rotatebox[origin=c]{90}{\#layers} & \rotatebox[origin=c]{90}{\makecell{\#neurons}} & \rotatebox[origin=c]{90}{activation} & \rotatebox[origin=c]{90}{\makecell{dropout\\factor}} & \rotatebox[origin=c]{90}{\#layers} & \rotatebox[origin=c]{90}{\makecell{filter\\dimension}} & \rotatebox[origin=c]{90}{\#filters} & \rotatebox[origin=c]{90}{activation} & \\ \hline
\multirow{3}{*}{VGG-16} & CalTech-101                        & 2 & [512, 256] & [ELU, SELU] & [0.3, 0.4] & - & - & - & - & 92.66 \\   
                              & CalTech-256                        & 1 & 1024 & SELU & 0.5 & - & - & - & - & 81.47 \\  
                              & Stanford Dogs                      & 1 & 512 & ELU & 0.4 & - & - & - & - & 81.65 \\ \hline
\multirow{3}{*}{ResNet-50} & CalTech-101                        & - & - & - & - & 3 & $[5\times5, 2\times2, 2\times2]$ & [512, 512, 256] & [SELU, SELU, SELU] & 94.81\\ 
                              & CalTech-256                        & - & - & - & - & 2 & $[2\times2, 3\times3]$ & $[512, 512]$ & [RELU, RELU] & 83.6\\  
                              & Stanford Dogs                      & - & - & - & - & 1 & $5\times5$ & 512 & Sigmoid & 81.43\\ \hline
\multirow{3}{*}{DenseNet-121} & CalTech-101                        & - & - & - & - & 2 & $[5\times5, 2\times2]$ & [512, 128] & [ELU, ELU] & 96.25\\ 
                              & CalTech-256                        & - & - & - & - & 2 & $[5\times5, 2\times2]$ & $[512, 512]$ & [SELU, RELU] & 84.35 \\ 
                              & Stanford Dogs                        & - & - & - & - & 2 & $[2\times2, 2\times2]$ & $[256, 512]$ & [RELU, RELU] & 82.21\\ \hline
\end{tabular}
}
\caption{The optimal hyperparameter's configuration of CNN layers learned for better transfer learning using the proposed Efficient AutoTune framework which uses Bayesian Optimization for architecture search. The pre-trained CNN's layers are tuned automatically w.r.t. the target datasets CalTech-101, CalTech-256, and Stanford Dogs. Layers like pooling have no suggested hyperparameter changes for VGG-16, ResNet-50 and DenseNet-121.}
\label{tab:results_proxy_Bayesian}
\end{table}

\begin{table}[t]
\centering
\resizebox{\textwidth}{!}{%
\begin{tabular}{|c|c|c|c|c|c|c|c|c|c|c|c|c|c|c|}
\hline
\multirow{2}{*}{Deep CNN}     & \multirow{2}{*}{\rotatebox[origin=c]{90}{Dataset}}          & \multicolumn{4}{c|}{\rotatebox[origin=c]{90}{FC Layer}}                                                                                                                                                & \multicolumn{4}{c|}{\rotatebox[origin=c]{90}{Convolution Layer}}                                                                                             &  \multirow{2}{*}{\makecell{Validation \\ Accuracy}} \\ 
\hhline{|~||~|-----------|}
                              &                                    & \#layers               & \begin{tabular}[c]{@{}c@{}} \#neurons \end{tabular} & activation            & \begin{tabular}[c]{@{}c@{}}dropout \\ factor\end{tabular} & \#layers               & \begin{tabular}[c]{@{}c@{}}filter\\ dimension\end{tabular} & \#filters              & activation   &                                                                                         \\ \hline
\multirow{1}{*}{CNN-static}       & \makecell{Moview Reviews \\ Sentiment Analysis}                       & 2                     & [64, 512]                                                        & [RELU, RELU]           & [0.9, 0.4]                                                & -                     & -                                                          & -                    & -                                                                              & 84.77                                                          \\ \hline

\end{tabular}
}
\caption{The optimal hyperparameter's configuration of CNN layers learned for better transfer learning using the proposed Efficient AutoTune framework which uses Bayesian Optimization for architecture search.}
\label{tab:results_proxy_Bayesian_NLP}
\end{table}

\begin{table*}
\centering 
\resizebox{\textwidth}{!}{%
\begin{tabular}{|c|c|l|c|c|c|c|}
\hline
\textbf{Dataset} &
  \textbf{Model} &
  \multicolumn{1}{c|}{\textbf{Method}} &
  \textbf{\begin{tabular}[c]{@{}c@{}}Accuracy\\ (in \%)\end{tabular}} & \textbf{\begin{tabular}[c]{@{}c@{}}Total Parameters \\ ($\times$ 10^6)\end{tabular}} &
  \textbf{\begin{tabular}[c]{@{}c@{}}Trainable Parameters\\ ($\times$ 10^6)\end{tabular}} &
  \textbf{\begin{tabular}[c]{@{}c@{}}Remaining FLOPs\\ ($\times$ 10^10)\end{tabular}} \\ \hline
\multirow{16}{*}{CalTech-101}   & \multirow{6}{*}{VGG-16}      & 
Traditional Transfer Learning & 86.65 & 134.6 & 14.45 & 204.43 (0\%)   \\   
                                &                              &
FCC \cite{qian2019single} & 91.77 & 134.6 & 14.45 & 204.43 (0\%)     \\  
                                &                              &

Shah \etal \cite{shah2020deriving}       & 87.3  & 117.5 & - & -     \\ 
                                &                              & AutoFCL \cite{basha2021autofcl}                    & 92.72 & 14.715 & 14.45 & 204.43 (0\%)     \\ 
                                &                              & AutoTune \cite{basha2021autotune}                     & 93.16 & 27.718 & 14.45 & 204.43 (0\%)     \\  
                                &                              & TASCNet-1                     & \textbf{94.15} & 25.93 &13.19 & 187.72 (8.17\%)  \\ 
                                &                              & TASCNet-2                     & 93.43 & \textbf{14.25} & \textbf{5.67}  & \textbf{87.42} \textbf{(57.24\%)}  \\ 
                                \hhline{|~|------|}
                                & \multirow{5}{*}{ResNet-50}    & Traditional Transfer Learning & 87.43 & 23.7 &  5.25  & 2.77 (0\%)       \\ 
                                &                              & 
                                FCC \cite{qian2019single} & \textbf{92.98} & 23.7 &  5.25  & 2.77 (0\%)       \\ 
                                &                              &
                                AutoFCL  \cite{basha2021autofcl}                     & 90.15 & \textbf{23.481} & 5.25  & 2.52 (9.03\%)    \\  
                                &                              & AutoTune \cite{basha2021autotune}                     & 92.0 & 26.561 & 4.21   & 2.52 (9.03\%)    \\  
                                &                              & TASCNet-1                     & 89.58 & 25.97 & 3.99  & 2.39 (13.72\%)   \\ 
                                &                              & TASCNet-2                     & 89.25 & 25.97 & \textbf{3.99}  & \textbf{2.39} \textbf{(13.72\%)}   \\ \hhline{|~|------|} 
                                & \multirow{5}{*}{DenseNet-121} & Traditional Transfer Learning & 82.8  & 7.05 & \textbf{0.66}  & \textbf{10.07 (0\%)}      \\  
                                &                              & AutoFCL \cite{basha2021autofcl}                       & 90.21 & \textbf{6.87} & 0.66  & 10.09 (-0.20\%)  \\ 
                                &                              & AutoTune \cite{basha2021autotune}                     & 91.14 & 20.087 & 13.35 & 10.09 (-0.20\%)  \\  
                                &                              & TASCNet-1                     & \textbf{95.8} & 19.08 & 12.43 & 9.66 (4.07\%)    \\ 
                                &                              & TASCNet-2                     & 94.79 & 16.40 & 10.01 & 8.33 (17.28\%)   \\ \hline
\multirow{16}{*}{CalTech-256}   & \multirow{6}{*}{VGG-16}      & Traditional Transfer Learning & 76.6  & 135.3 & 14.45 & 204.43 (0\%)     \\ 
                                &                              &
                                FCC \cite{qian2019single} & \textbf{84.61} & 135.3 &  14.45  & 204.43 (0\%)       \\  
                                &                              &
                                 Shah \etal \cite{shah2020deriving}       & 77.5  & 118.1 & - & -     \\
                                &    & 
               PSBWN \cite{gadosey2019pruned}                       & 65.4 & 59 & - & 171.72 (16\%)     \\  
                                &                              &

                                AutoFCL \cite{basha2021autofcl}                       & 77.19 & \textbf{27.692} & 14.45 & 204.43 (0\%)     \\ 
                                &                              & AutoTune     \cite{basha2021autotune}                 & 80.92 & 119.709 & 14.45 & 204.43 (0\%)     \\  
                                &                              &
                                TransTailor  \cite{liu2021transtailor}                    & 81.8 & - & - & 163.54 (20\%)     \\ 
                                & &
                                TASCNet-1                     & 82.21 & 114.63 & 13.18 & 187.72 (8.17\%)  \\  
                                &                              & TASCNet-2                     & 81.19 & 108.86 & \textbf{12.03} & \textbf{173.39 (15.18\%)} \\ \hhline{|~|------|}
                                & \multirow{5}{*}{ResNet-50}    & Traditional Transfer Learning & 63.24 & \textbf{24.05} & 5.25  & 2.77 (0\%)       \\ 
                                &                              &
               FCC \cite{qian2019single} & 82.78 & 24.05 &  5.25  & 2.77 (0\%)       \\  
                                &                              &                 AutoFCL \cite{basha2021autofcl}                   & 80.48 & 25.843 & 5.25  & 2.77 (0\%)       \\  
                                &                              & AutoTune \cite{basha2021autotune}                     & 81.99 & 25.843 & 5.25  & 2.77 (0\%)       \\  
                                &                              & TASCNet-1                     & 81.42 & 25.37 & 4.98  & 2.63 (5.05\%)    \\  
                                &                              & TASCNet-2                     & 81.03 & 24.50 & \textbf{4.49}   & \textbf{2.38 (14.08\%)}   \\ 
                                \hhline{|~|------|}
                                & \multirow{5}{*}{DenseNet-121} & Traditional Transfer Learning & 63 & \textbf{7.2}     & 0.66  & 10.07 (0\%)      \\  
                                &                              & AutoFCL  \cite{basha2021autofcl}                       & 81.6  & 8.183 & 0.66  & 10.07 (0\%)      \\  
                                &                              & AutoTune  \cite{basha2021autotune}                    & 83.43 & 8.183 & 0.66  & 10.07 (0\%)      \\  
                                &                              & TASCNet-1                     & \textbf{83.74} & 8.04 & 0.62  & 9.64 (4.27\%)    \\ 
                                &                              & TASCNet-2                     & 82.04 & 7.77 & \textbf{0.55}  & \textbf{8.77 (12.91\%)}   \\ \hline
\multirow{15}{*}{Stanford Dogs} & \multirow{5}{*}{VGG-16}      & Traditional Transfer Learning & 78.57 & 134.7 &  14.45 & 204.43 (0\%)     \\ 
                                &                              & AutoFCL \cite{basha2021autofcl}                      & 72.08 & \textbf{17.942} & 14.45 & 204.43 (0\%)     \\  
                                &                              & AutoTune     \cite{basha2021autotune}                 & 81.15 & 119.638 & 14.45 & 204.43 (0\%)     \\  
                                &                              &
                                TransTailor  \cite{liu2021transtailor}                    & 82.6 & - & - & 163.544 (20\%)     \\  
                                &                              &
                                TASCNet-1                     & \textbf{82.7} & 114.55 & 13.18 & 187.72 (8.17\%)  \\  
                                &                              & TASCNet-2                   & 81.82 & 103.05 &\textbf{10.91} & \textbf{157.4 (23.00\%)}  \\ 
                                \hhline{|~|------|}
                                & \multirow{5}{*}{ResNet-50}    & Traditional Transfer Learning & 81.26 & \textbf{23.7} &  5.25  & 2.77 (0\%)       \\  
                                &                              & AutoFCL  \cite{basha2021autofcl}                     & 78.93 & 25.179 & 5.25  & 2.77 (0\%)       \\ 
                                &                              & AutoTune   \cite{basha2021autotune}                   & 80.49  & 25.703 & 5.25  & 2.77 (0\%)       \\  
                                &                              &
                                DTQ \cite{xie2020deep}                      & 82.3  & - & -  & 2.77 (0\%)       \\ 
                                &                              &
                                TASCNet-1                     & \textbf{81.68} & 25.70 & 5.24  & 2.76 (0.36\%)    \\  
                                &                              & TASCNet-2                     & 78.06 & 25.23 & \textbf{4.98}  & \textbf{2.63 (5.05\%)}    \\ \hhline{|~|------|} 
                                & \multirow{5}{*}{DenseNet-121} & Traditional Transfer Learning & \textbf{82.5}  & \textbf{7.07} & 0.66  & 10.07 (0\%)      \\ 
                                &                              & AutoFCL \cite{basha2021autofcl}                      & 81.24 & 7.457 & 0.66  & 10.07 (0\%)      \\ 
                                &                              & AutoTune     \cite{basha2021autotune}                 & 81.55 & 7.457 & 0.66  & 10.07 (0\%)      \\  
                                &                              & TASCNet-1                     & 82.28 & 7.46 & 0.655 & 9.67 (3.97\%)\\    
                                &                              & TASCNet-2                     & 78.81 & 7.33 &\textbf{0.62}  & \textbf{9.64 (4.27\%)}\\ \hline

\multirow{3}{*}{\makecell{Movie Reviews \\  Sentiment Analysis}} & \multirow{3}{*}{CNN-static}      & Traditional Transfer Learning & 83.00 & 36.07 &  1.23 & 0.298 (0\%)     \\ 
                                 
                                &                              &
                                TASCNet-1                     & \textbf{86.12} & 0.99  & 0.89 &  0.215(27.53\%)  \\  
                                &                              & TASCNet-2                   &  85.76 & \textbf{0.86} & \textbf{0.77} & \textbf{ 0.186(37.30\%)}  \\
                                 \hline
\end{tabular}
}
\caption{Comparison of transfer learning results obtained using TASCNet with state-of-the-art methods in terms of classification accuracy, number of trainable parameters, and number of remaining FLOPs. Here, TascNet-1, TascNet-2 entries correspond to the top performing results in terms of accuracy and FLOPs reduction, respectively.} \label{table:4MethodSummary}
\end{table*}

\section{Experimental Settings}
\label{exps_results}
This section describes the settings, such as the training details, CNNs used for tuning and pruning w.r.t. target dataset, and the datasets used to conduct the experiments.

\subsection{Training Details}
\label{training_details}
\textbf{Training Proxy (Child) CNNs:}
The pre-trained CNN layers explored during the hyperparameter tuning in the first stage are trained using the Adagrad optimizer \cite{duchi2011adaptive} for 50 epochs with the learning rate as $1.0\times10^{-2}$ using cross-entropy loss for image classification. Later, we decrease the learning rate by a factor of $\sqrt{0.1}$ upon observing the degradation in the validation loss. 

\textbf{Training details while Pruning CNNs:} From the target-aware pre-trained CNNs obtained in the first phase, filters are pruned to decrease the number of parameters and FLOPS. From VGG-16 and DenseNet-121, we prune filters from the convolutional layers having \#filters greater than or equal to 256 to avoid possible loss of information on pruning layers having fewer \#filters. For ResNet-50, we prune filters from the convolutional layers having \#filters greater than or equal to 2048 due to the steep decrease in validation accuracy which we have observed empirically. Each pruning iteration involves re-training (optimizing) the model for 50 epochs with the custom loss function before pruning and retraining (finetuning) the model for 50 epochs after pruning the filters. The pruning process terminates when the difference between the maximum validation accuracy and the current iteration validation accuracy exceeds a threshold. In our experiments, this threshold was set to 0.02 for all the models. 


\subsection{Deep CNNs utilized for Tuning and Pruning}
\label{CNNs_tuning_and_pruning}
The proposed deep ConvNets from the literature can be classified into two kinds: viz, i) Chain-structured CNNs
ii) CNNs containing skip-connections
\\
\textbf{Chain-structured (Plain) CNNs:} The CNNs proposed during the early stage of deep CNNs \cite{lecun1998gradient, krizhevsky2012imagenet} follow a chain-structured hierarchy in which $n$ layers are stacked horizontally such that $i^{th}$ layer accepts the input from layer $i-1$ and its output feature-map fed as input to layer $i+1$. 
In this paper, we utilize VGG-16 to automatically tune the pre-trained CNN w.r.t. a target dataset, thereby pruning the redundant filters to increase the network's generalization ability. 
\\
\textbf{CNNs containing Skip Connections:}
Another class of CNNs have skip connections that allow a layer to accept input feature maps from more than one of their previous layers. For example, let us assume a CNN with n layers which is formally illustrated as \{$L_1, L_2, ...., L_{n-1}, L_n$\}. Here $L_1, L_{n}$ be the input and the output layers of a CNN, respectively.
To automatically tune the pre-trained CNNs and later prune the redundant filters, we used ResNet-50 and DenseNet-121.

\subsection{Datasets}
To justify the significance of the proposed method, we utilize three publicly available image datasets CalTech-101 \cite{fei2004learning}, CalTech-256 \cite{griffin2007caltech}, and Stanford Dogs \cite{khosla2011novel}. The higher-level details of these datasets are shown in Table \ref{table:datasetsummary}. The 20 News Group \cite{lang1995newsweeder}, Movie Reviews Sentiment Analysis \cite{maas2011learning} datasets are used as source and target datasets, respectively, to verify the robustness of the proposed method for NLP tasks.

\section{Results and Discussions}
\label{sec:results_discussion}
The first stage proposed framework involves automatically tuning the hyperparameters involved in the deeper layers of a pre-trained CNN for a target dataset. We utilize the Bayesian Optimization \cite{frazier2018tutorial} to tune the hyperparameters using the search space presented in Table  \ref{hyperparameter_space_table}. The procedure of this stage is described in Algorithm \ref{Eff_Auto_Tune_algorithm}.

We have conducted experiments using VGG-16, ResNet50, and DenseNet121 on CalTech-101, CalTech-256, and Stanford Dogs datasets. The resulting structures of deeper layers after hyperparameter tuning are shown in Table \ref{tab:results_proxy_Bayesian}. The original VGG-16 architecture contains 2 fully connected layers (excluding the output layer) each with 4096 neurons. Conventionally, the parameters corresponding to the last layer or last few layers are updated using the target dataset. We obtain 86.65\% classification accuracy on CalTech-101 using traditional transfer learning. This is achieved by fine-tuning the target-independent CNN layers. After employing hyperparameter tuning for VGG-16 in the first stage, we observe that with 2 fully connected (FC) layers, one FC layer with 512 neurons, ELU activation, dropout rate of 0.3 and another FC layer with 256 neurons, SELU activation, dropout rate of 0.4 has achieved an improved accuracy of 92.66\%. In the case of ResNet-50 and DenseNet-121, we can also observe the architectural changes in the convolutional layers.

The CNNs listed in Table \ref{tab:results_proxy_Bayesian} are designed to achieve SOTA performance on the ImageNet dataset. While the hyperparameter tuning stage of TASCNet has made some architectural changes, those changes are mostly limited to the fully connected layers or convolutional layers in some cases. In other words, the hyperparameters involved in the last few layers of a pre-trained CNN are learned using the target dataset. However, most of the other layers are designed for source dataset that may not be relevant to the target dataset. To address this problem, we prune the redundant filters from convolution layers in the second stage of TASCNet viz. the pruning stage.

In Table \ref{table:4MethodSummary}, we present a performance comparison of the proposed TASCNet method with state-of-the-art, including conventional transfer learning. We compare the proposed method results with the existing methods w.r.t. classification accuracy, the number of trainable parameters, and remaining FLOPS in the convolution layers from which the filters are pruned. We observe that over the CalTech-101 and CalTech-256 datasets, TASCNet performs better than the existing methods for VGG-16 and DenseNet-121. Whereas, for ResNet-50, our method achieves comparable results with AutoTune \cite{basha2021autotune} on CalTech-101 and CalTech-256. However, for the fine-grained classification dataset, i.e., Stanford Dogs, the proposed TASCNet performs better in terms of validation accuracy and reduced total FLOPS, compared to other methods for both VGG-16 and ResNet-50. For DenseNet-121, traditional transfer learning performs better.

To verify the efficacy of the proposed TASCNet method for NLP tasks, experiments are conducted for Movie Review Sentiment Analysis. Glove embeddings \cite{pennington2014glove} are extracted for both the source and target datasets, which are considered as input to CNNs. The CNN-static model \cite{kim2014convolutional} is trained over ``The 20 News Group" dataset \cite{lang1995newsweeder}, which is considered as the pre-trained model for further experiments. All layers of pre-trained CNN-static are fine-tuned on the Movie Review Sentiment Analysis dataset \cite{maas2011learning} except the embedding layer. Through this experiment,  83\% accuracy is observed on the validation set. To compare the proposed method results with traditional transfer learning, hyperparameters of the CNN-static \cite{kim2014convolutional} are tuned w.r.t. target dataset, i.e., Movie Reviews Sentiment Analysis. The optimal hyperparameters learned are presented in Table \ref{tab:results_proxy_Bayesian_NLP}.  Later, the model with the learned hyperparameters is fine-tuned over the Movie Reviews Sentiment Analysis dataset. From table \ref{tab:results_proxy_Bayesian_NLP}, we can observe that tuning the hyperparameters w.r.t. the target dataset leads to improved results (84.77\%) compared to conventional fine-tuning. Finally, the fine-tuned model is considered as the pre-trained model to conduct pruning experiments. The TASCNet results of Movie Reviews Sentiment Analysis are compared with traditional fine-tuning in Table \ref{table:4MethodSummary}. It is evident from this table that TASCNet produces improved results compared to traditional fine-tuning with less storage (trainable parameters) and compute resources (FLOPs).       

\subsection{Advantages \& Disadvantages of TASCNet}

TASCNet’s superior performance, while having less number of FLOPS and trainable parameters than other methods, can be attributed to i) the efficient hyperparameter optimization over both convolutional layers and dense layers of a pre-trained CNN, ii) pruning the redundant filters. This enables better generalization ability of the CNN and ensures that the remaining filters take an active part in learning features essential for the target task. From Table \ref{table:4MethodSummary}, we can note that using the TASCNet method, better transfer results are achieved with fewer computations and memory requirements. However, the proposed method involves two-stage process to find optimal pre-trained CNN configuration for efficient transfer learning, which may be achieved in single stage in the near future.

\section{Conclusion and Future Scope}
\label{conclusion}
This paper proposes a two-stage framework to find a lightweight pre-trained CNN for a given target dataset that allows resource-efficient knowledge inference. In the first stage, the hyperparameters involved in the top layers of a pre-trained CNN are automatically tuned with the knowledge of the target dataset. Later, in the second stage, we prune the filters that are redundant throughout training by pairing them based on cosine similarity. Using the proposed TASCNet method, we can reduce both trainable parameters and FLOPs required for transfer learning. Our empirical results show that the pre-trained CNN's transfer learning ability is improved using this two-stage framework. We believe that these methods are best suited for applications where quick inference plays a vital role. Pruning the less significant filters for the target task and fine-tuning specific filters may be a potential research direction. The proposed method can be extended by combining the current two stage approach to single stage so that human intervention can be minimized.  

\section*{Acknowledgments}

We acknowledge the NVIDIA Corporation's support with the donation of GeForce Titan XP GPU, which is helpful in conducting the experiments of this research.

\section{Conflicts of Interest}
We declare the following conflicts of interest,
\begin{enumerate}
    
    \item Indian Institute of Information Technology Sri City, Chittoor, India.
    \item Lytx India Technologies Private Limited, Bangalore.
    \item RV University Bangalore, India.
    \item Tata Consultancy Services, Hyderabad, India.
    \item New Jersey Institute of Technology, New Jersey, USA.
    \item Google Research Bangalore, India.
    \item Indian Institute of Information Technology Allahabad, India.
\end{enumerate}

\bibliography{mybibfile}

\end{document}